\documentclass[]{fairmeta}
\usepackage[utf8]{inputenc}
\usepackage{amsmath}
\usepackage{amssymb}

\usepackage{array}
\usepackage{tabularx}
\usepackage{colortbl}
\usepackage{wrapfig}
\usepackage{adjustbox}
\usepackage{placeins}
\usepackage{flafter}

\usepackage{xspace}

\providecommand{\keywords}[1]{}
\title{AffordanceWAM: Affordance-Aware Joint World--Action Modeling for Robot Manipulation}

\author[1,3,\dagger]{Jiadi You}
\author[2,\dagger]{Qize Yu}
\author[2]{Yue Chen}
\author[4]{Minghong Cai}
\author[3]{Zhide Zhong}
\author[2,5]{Yuran Wang}
\author[2]{Bowen Ping}
\author[2]{Jiaqi Liang}
\author[2]{Zhenhao Shen}
\author[3]{Haodong Yan}
\author[6]{Yinchuan Li}
\author[2]{Ruihai Wu}
\author[1,*]{Xiaojuan Qi}
\author[3,*]{Yingcong Chen}

\affiliation[1]{The University of Hong Kong}
\affiliation[2]{Peking University}
\affiliation[3]{HKUST (GZ)}
\affiliation[4]{The Chinese University of Hong Kong}
\affiliation[5]{National University of Singapore}
\affiliation[6]{Knowin AI}

\contribution[\dagger]{Equal contribution.\quad${}^{*}$Corresponding authors.}

\abstract{
Generalizable robot manipulation requires predicting how a scene will evolve,
identifying where interactions are feasible, and determining how to act.
Action-labeled robot videos directly supervise control but are costly and
limited in diversity, whereas egocentric human videos capture diverse
interactions but lack robot actions and differ in embodiment and appearance.
We introduce AffordanceWAM, an affordance-aware generative World Action Model
that represents object-centric spatiotemporal affordance through Scalar
Affordance and Affordance Heatmap, within the generated future World.
This representation grounds visual prediction in task-relevant objects and
interaction regions for action generation, and provides shared interaction targets across
human and robot videos. Built on a pretrained video diffusion Transformer,
AffordanceWAM uses separately parameterized World and Action Experts, coupled
through Masked Joint Self-Attention, to jointly predict future RGB
observations, Scalar Affordance fields, Affordance Heatmaps, and continuous
robot actions under a unified flow-matching objective. Human videos supervise
all three future-World streams, whereas robot trajectories additionally provide action
supervision, enabling transfer without human action labels or retargeting.
Experiments on RoboCasa, CALVIN ABC$\rightarrow$D, and real-world manipulation
demonstrate consistent gains over RGB-only and robot-data-only baselines. Under
fixed robot supervision, RoboCasa performance improves monotonically as
affordance-annotated human video scales. These results support affordance as an
effective interface for both vision--language--action learning and
human-to-robot transfer.
}

\metadata[Project Page]{\url{https://skywalker-yqz.github.io/AffordanceWAM/}}
\metadata[Code]{\url{https://github.com/Skywalker-yqz/AffordanceWAM}}

\begin{document}

\maketitle

% ---------------------------------------------------------------
% Main content
\section{Introduction}
\label{sec:introduction}

Egocentric human data offers a source for scaling robot
manipulation. Humans perform dexterous manipulation across diverse objects,
environments, and tasks, which is difficult to match through teleoperation
pipelines \cite{jain2024vid2robot,kareer2025egomimic,chen2025vidbot,wang2026humanego}. Recent work shows that this experience can improve robot
learning by aligning human demonstrations with robot actions \cite{kareer2025emergence,kareer2025egomimic,wang2026humanego,zheng2026egoscale,punamiya2025egobridge}. However, these gains depend on alignment in viewpoint, speed,
and behavior. Without this bridge, human data may degrade performance.

Recently, World Action Models (WAMs) provide a more direct alternative. Alongside action
prediction, they predict future states from a shared backbone \cite{uva,uwm,cen2025worldvla,li2026lingbotva,ye2026dreamzero}.
Human and robot videos can therefore supervise object and scene dynamics, which can be transferred across embodiments more readily than actions. By separating how the world evolves from embodiment-specific action execution, WAMs can use large-scale human data more effectively, yet appearance-oriented World representations remain semantically misaligned with action generation.

The question is which World representation supports transfer across
embodiments during WAM co-training. Most WAMs reconstruct RGB frames~\cite{yuan2026fast,guo2026unified} or visual latents~\cite{ye2025latent,luo2026jointaligned,zheng2025flare}, modeling appearance, background motion, and viewpoint changes while leaving interaction structure implicit. Affordance~\cite{gibson1977theory} makes interaction structure shared across embodiments explicit by identifying relevant objects~\cite{yu2026groundanythingreconcilingparalleldecoding,yu2026groundingpigroundingfoundationmodel} and feasible interaction regions and capturing how the task-relevant state evolves~\cite{bahl2023vrb,nasiriany2024rtaffordance,yuan2025robopoint,xu2025affordancefield,A3D}.
By encoding object-centric interaction semantics rather than embodiment-specific motion, affordance both bridges human and robot data and provides a semantic interface from World prediction to action generation.
% Because it describes interaction rather than embodiment-specific motion, affordance can supervise both human and robot data.

\begin{figure*}[t]
\centering
\includegraphics[width=\textwidth]{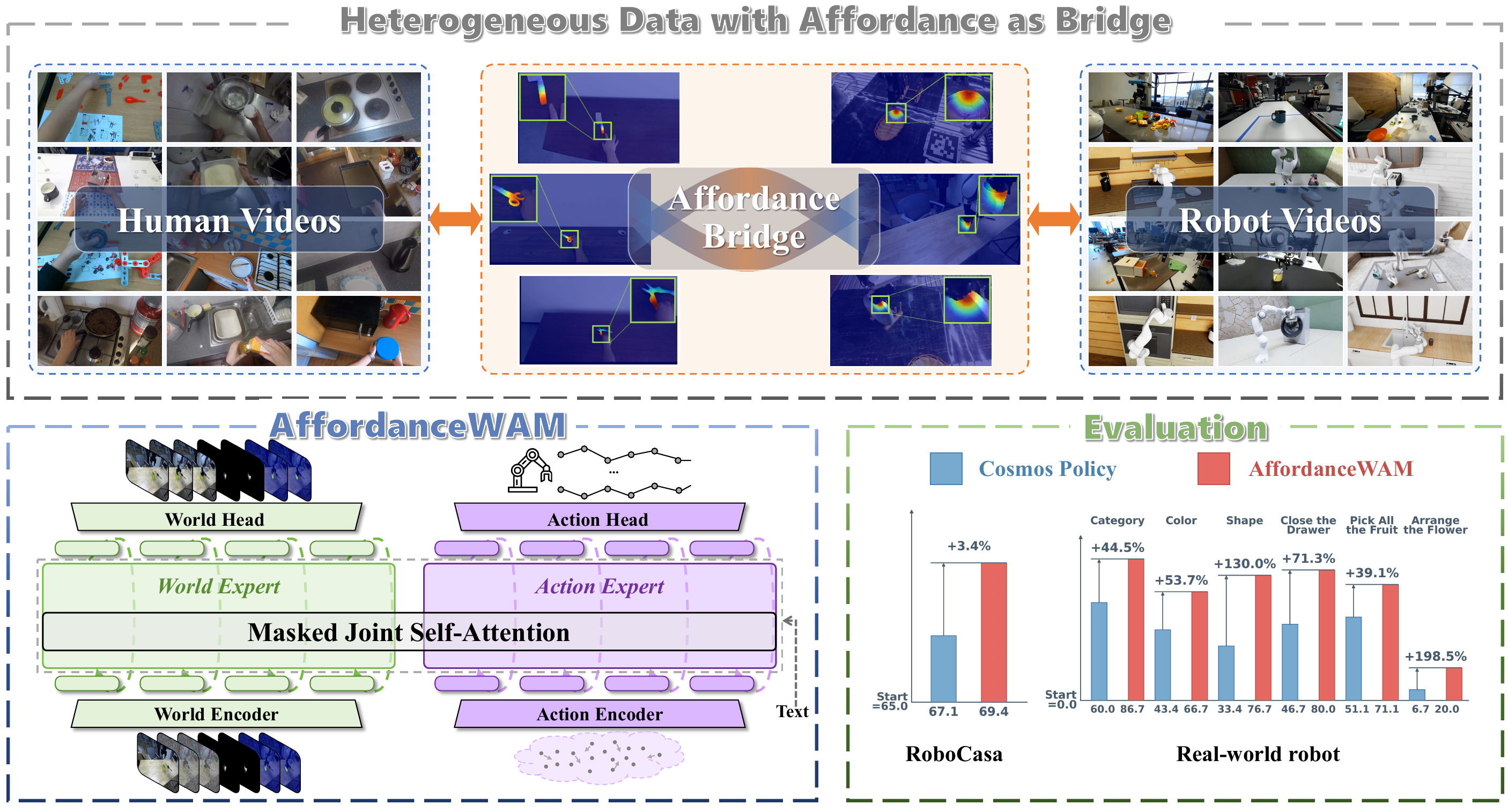}
\caption{AffordanceWAM overview.
A shared object-focused affordance representation bridges heterogeneous
human--robot data with joint world and action modeling, enabling positive
transfer to robot manipulation.}
\label{fig:teaser}
\end{figure*}

We therefore introduce \textbf{AffordanceWAM}, a dual-expert WAM designed to
share interaction dynamics across embodiments (Figure~\ref{fig:teaser}).
A pretrained video DiT serves as the World Expert, forecasting future RGB
observations and two complementary affordance components: Scalar Affordance
and Affordance Heatmap. An Action
Expert predicts continuous controls from the layer-wise RGB and Scalar Affordance
features of the evolving World.
To preserve this separation, Masked Joint Self-Attention lets the Action Expert
use these evolving World features, but prevents actions from influencing World
prediction and excludes the auxiliary Affordance Heatmap features from the
Action input. This reduces an additional RGB-rich pathway and encourages
greater use of Scalar Affordance while retaining access to RGB features. This directional
coupling leads to asymmetric training: human clips update only the shared World
Expert, whereas robot trajectories additionally train the Action Expert. Human
and robot data thus meet in interaction space rather than action space.

Experiments on RoboCasa~\cite{robocasa} and CALVIN~\cite{calvin} show strong
performance across tasks and unseen environments. Crucially,
% controlled ablations show that affordance determines whether human data helps.
controlled ablations show that affordance is the key bridge through which human interaction dynamics benefit robot learning.
Without affordance, human videos lower RoboCasa success from $57.3\%$
to $55.1\%$ and CALVIN sequence length from $3.75$ to $3.68$; with affordance,
the same data raises these metrics from $63.7\%$ to $69.4\%$ and from $3.91$ to
$4.22$. These gains carry over to real-world manipulation. 

In summary, our key contributions are summarized as follows:
\begin{itemize}
    \item We introduce AffordanceWAM, a WAM architecture that uses structured affordance forecasting as intermediate supervision to bridge predictive world modeling and robot control enabling generalization across diverse tasks.
    % We introduce AffordanceWAM, a WAM architecture that leverages structured affordance forecasting as intermediate supervision enabling generalization across diverse tasks.
    \item Evidence that affordance forecasting unlocks human-data scale by separating shared interaction dynamics from embodiment-specific control.
    \item Across simulation and real-world experiments, AffordanceWAM outperforms strong VLAs and WAMs demonstrating generalization and robustness across ablations, qualitative and quantitative analyses.
    % We achieve strong performance in both simulation and real-world experiments. AffordanceWAM demonstrates success rates competitive with recent strong VLAs and WAMs, strong generalization and robustness, supported by comprehensive ablation, qualitative and quantitative analyses.
\end{itemize}

\section{Related Work}
\label{sec:related-work}
\subsection{World Action Models}
\label{sec:world-action-models}

Video-based world models learn scene dynamics by predicting how visual
observations evolve over time. In robot learning, predicted videos can serve as
visual plans or provide predictive representations to a downstream
policy~\cite{du2023unipi,vpp}. These approaches connect visual prediction to
control, but the video model and action policy remain separate. World Action
Models instead learn future observations and actions within a unified
framework~\cite{song2026uniwamunifiedworldactionmodel}. Existing designs include coupled video--action diffusion
models~\cite{uva,uwm} and autoregressive models over visual and action
tokens~\cite{cen2025worldvla,li2026lingbotva}. More recent work builds on
pretrained video backbones or large-scale causal video--action
pretraining~\cite{ye2026dreamzero,zhang2026lingbotva2,wang2026openwamopenmodularexploration}. Joint modeling allows
visual dynamics to inform action learning and provides a route for using videos
without action annotations. Across these formulations, visual prediction may
act as a planning space, an auxiliary learning signal, or a jointly generated
output.

Despite their architectural differences, these methods usually represent the
predicted World with future RGB observations or visual latent tokens. Such
representations capture appearance and scene evolution, but they do not
explicitly identify the objects and interaction regions that determine a
manipulation task. AffordanceWAM augments the generated future World with
object-centric spatiotemporal Scalar Affordance and Affordance Heatmaps,
and jointly models these streams with future RGB observations and continuous
robot actions. It thereby makes
interaction structure an explicit prediction target rather than leaving it
implicit in visual features.

\subsection{Robot Learning from Human Videos}
\label{sec:human-video-learning}
Human videos have supported robot learning at increasingly direct levels.
Earlier methods pretrain visual representations or extract contacts, hand
trajectories, and object motion as planning or control interfaces
\cite{srirama2024hrp,bahl2023vrb,chen2025vidbot,wang2026humanego,yuan2024generalflow}.
Recent methods retarget human motion into robot actions and co-train policies across embodiments
\cite{jain2024vid2robot,kareer2025egomimic,zheng2026egoscale}. EgoScale shows that this route can benefit from large-scale egocentric data, but still relies
on retargeted wrist and hand actions followed by aligned human-robot
mid-training~\cite{zheng2026egoscale}. 
More broadly, action-level transfer
depends on the quality of human motion recovery and cross-embodiment alignment~\cite{wu2026sugarscalablehumanvideodrivengeneralizable}.

Future-state prediction offers a complementary route. Recently, WAMs couple future
prediction with action generation, allowing human videos to shape policy
representations through scene dynamics
\cite{uwm,cen2025worldvla,ye2026dreamzero, li2026egowam,shen2026ld4wamlearninglatentdynamics}. AffordanceWAM retains
future-state prediction but augments it with an affordance representation
comprising Scalar Affordance and Affordance Heatmap. Together, they identify
task-relevant interaction regions and associate their evolution with visual context.

% \noindent\textbf{Affordance Learning for Robotic Manipulation.}
\subsection{Affordance Learning for Robotic Manipulation}
\label{sec:affordance-learning}
Affordances describe the interaction possibilities offered by
objects~\cite{gibson1977theory}. Prior work represents them as contacts or
trajectories~\cite{bahl2023vrb,yuan2025robopoint}, structured
plans~\cite{nasiriany2024rtaffordance,xu2025a0}, or dense
fields~\cite{xu2025affordancefield,chen2026learning}. Recent affordance-aware VLAs further
incorporate these cues through reasoning, feature alignment, or structured
forecasting~\cite{li2025coavla,kong2026affordvla,yu2026affordancevla}, but
primarily use them as policy inputs or intermediate predictions. AffordanceWAM
instead treats temporally evolving Scalar Affordance and Affordance Heatmap
as parts of the generated future World. During joint pretraining, human and
robot videos share these prediction targets, while the Action Expert reads layer-wise hidden
features from the RGB and Scalar Affordance streams and receives action
supervision only from robot trajectories. Affordance therefore serves as a cross-embodiment
World representation rather than an embodiment-specific action target.

\section{Method}
\label{sec:method}

\begin{figure*}[t]
\centering
\includegraphics[width=\textwidth]{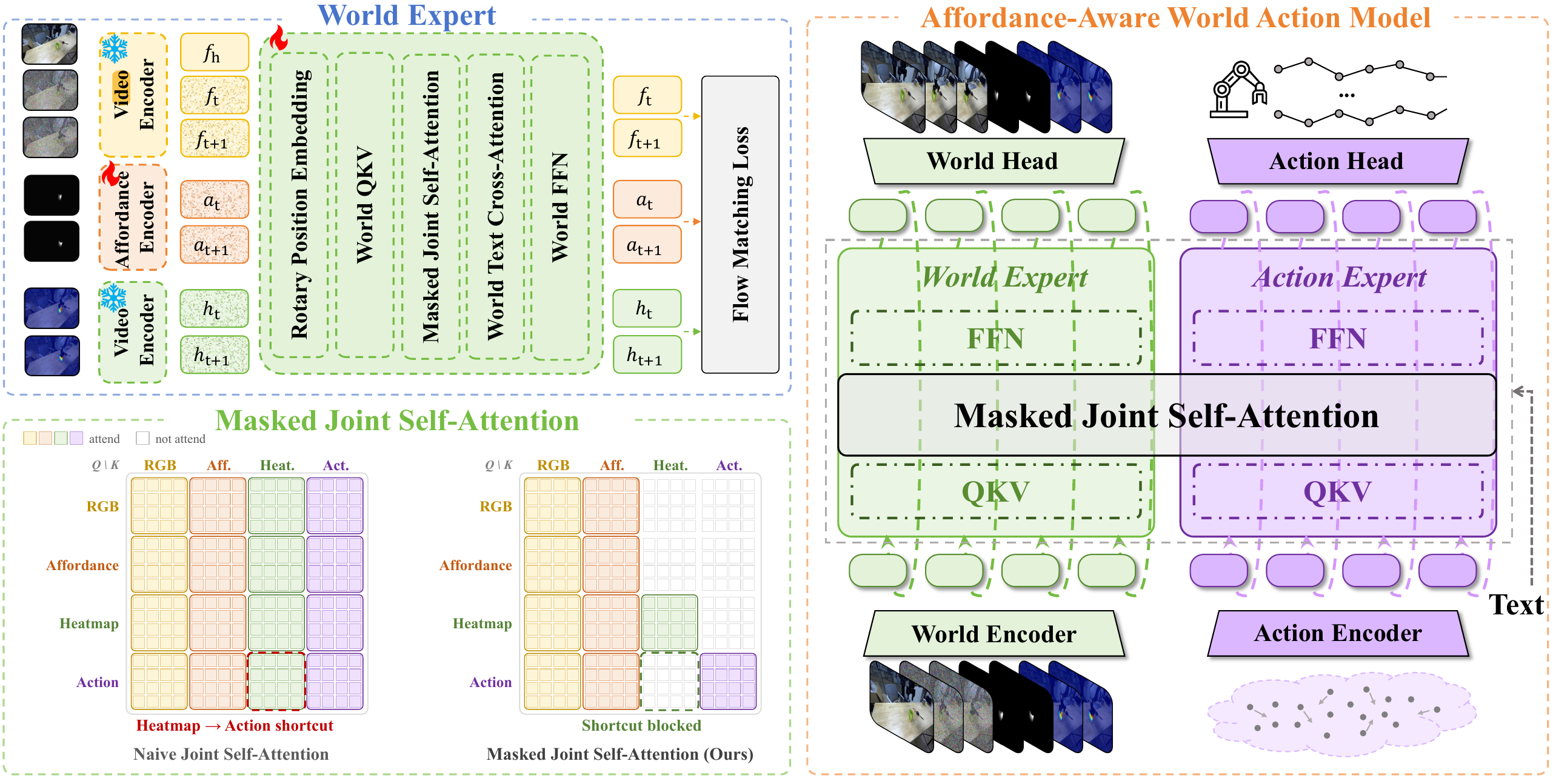}
\caption{AffordanceWAM overview. The World Expert jointly predicts future
RGB, Scalar Affordance, and Affordance Heatmaps. The Action Expert
reads RGB and Scalar Affordance hidden features through layer-wise Masked Joint
Self-Attention to generate continuous actions. The mask excludes Action
features from World prediction and Affordance Heatmap features from the
Action input, reducing an additional RGB-rich pathway to control.}
\label{fig:method_overview}
\end{figure*}

AffordanceWAM learns a shared interaction representation from human and
robot videos and grounds it in robot control. Its World Expert predicts
three future-World streams---RGB, Scalar Affordance, and Affordance Heatmap---while
its Action Expert generates actions by accessing the evolving RGB and
Scalar Affordance hidden features at each Transformer layer
(Fig.~\ref{fig:method_overview}). We
introduce the generative formulation, describe the architecture and World
representation, and then present learning from heterogeneous supervision.

\subsection{Preliminaries}
\label{sec:preliminaries}

\paragraph{Flow matching.}
Flow matching learns a velocity field that transports Gaussian noise to
data. Given a clean target $y_0$, noise
$\epsilon\sim\mathcal N(0,\mathrm{I})$, and flow time
$\tau\in[0,1]$, a linear interpolation defines the noisy input and target
velocity:
\begin{equation}
  y_\tau=(1-\tau)y_0+\tau\epsilon,
  \qquad u=\epsilon-y_0.
  \label{eq:flow_path}
\end{equation}
Given conditioning
information $c$, a velocity predictor $v_\theta$ with learnable parameters
$\theta$ is trained by
\begin{equation}
  \mathcal L_{\mathrm{FM}}
  =\mathbb E_{y_0,c,\epsilon,\tau}\!\left[
    \frac{1}{n}\|v_\theta(y_\tau,\tau;c)-(\epsilon-y_0)\|_2^2
  \right],
  \label{eq:stream_flow_loss}
\end{equation}
where $n$ is the number of scalar target entries and the norm is computed
over flattened entries. Generation starts from noise at $\tau=1$ and
integrates the learned velocity back to the data endpoint at $\tau=0$.
We apply this formulation to both future World latents and continuous
actions; subscripts $0$ and $\tau$ distinguish clean and noisy states.

\paragraph{World--action modeling.}
Let $I^-=(I_0,\ldots,I_{P-1})$ be a history of $P$ RGB frames, $\ell$ a
language instruction, and $s$ the current proprioceptive state. Action
modeling predicts a continuous chunk
$U_0\in\mathbb R^{T_{\mathrm A}\times d_{\mathrm A}}$ from this context,
where $T_{\mathrm A}$ and $d_{\mathrm A}$ denote the action horizon and
dimension. AffordanceWAM additionally defines the future-World target
$\mathcal W^+=(I^+,F^+,M^+)$ comprising $T$ future RGB frames
$I^+=(I_P,\ldots,I_{P+T-1})$, aligned Scalar Affordance $F^+$,
and Affordance Heatmaps $M^+$. The corresponding latent variables, together
with the action chunk, are modeled by
\begin{equation}
  p_\theta(\mathcal W^+,U_0\mid I^-,\ell,s).
  \label{eq:joint_world_action}
\end{equation}
The generated future World provides an explicit representation of scene evolution
and interaction regions for action generation. We parameterize this joint
model through coupled World and Action velocity predictors, with World
generation conditioned on RGB history and language and proprioception
provided to the Action Expert. During training, $I^+$, $F^+$, and $M^+$ are
clean supervision targets. At deployment, these clean targets are unavailable;
their latent variables are generated from the conditioning context. Frame
indices $t$ are distinct from flow time $\tau$.

\subsection{AffordanceWAM Architecture}
\label{sec:wam_architecture}

AffordanceWAM consists of a World Expert $\mathcal{M}_{\mathrm W}$ and
an Action Expert $\mathcal{M}_{\mathrm A}$. Both are diffusion Transformers
with the same number of layers, but have separate parameters and
modality-specific hidden dimensions. Their coupling occurs through joint
attention at corresponding layers, allowing the Action Expert to use evolving
World hidden features throughout joint denoising.

\subsubsection{World Expert.}

The World Expert is initialized from a pretrained video diffusion
Transformer. Let $C$ denote the encoded RGB history and
$Z_0^m$ the clean future latent for stream
$m\in\mathcal{S}_{\mathrm W}=\{\mathrm{rgb},\mathrm{afd},\mathrm{heat}\}$;
the encoders are defined in Sec.~\ref{sec:world_representation}.
The indices $\mathrm{afd}$ and $\mathrm{heat}$ identify Scalar Affordance
and Affordance Heatmap, respectively.
At World flow time $\tau_{\mathrm W}$, collect the three noisy future
latents into $\mathbf Z_{\mathrm W,\tau_{\mathrm W}}
=(Z_{\tau_{\mathrm W}}^{\mathrm{rgb}},
Z_{\tau_{\mathrm W}}^{\mathrm{afd}},
Z_{\tau_{\mathrm W}}^{\mathrm{heat}})$. Conditioned on RGB history and
language, the Expert predicts their flow velocities:
\begin{equation}
  \widehat{\mathbf v}_{\mathrm W}
  =\mathcal M_{\mathrm W}
  \bigl(\mathbf Z_{\mathrm W,\tau_{\mathrm W}},\tau_{\mathrm W};C,\ell\bigr).
  \label{eq:world_expert}
\end{equation}
Here $\widehat{\mathbf v}_{\mathrm W}
=(\widehat v_{\mathrm W}^{\mathrm{rgb}},
\widehat v_{\mathrm W}^{\mathrm{afd}},
\widehat v_{\mathrm W}^{\mathrm{heat}})$ is the tuple of predicted velocities.
For $m\in\mathcal S_{\mathrm W}$, $\widehat v_{\mathrm W}^{m}$ denotes
its $m$-th stream component and has the same shape as $Z_0^m$.
The hat denotes a predicted velocity, the subscript identifies the Expert,
and the superscript identifies a World stream. History remains clean and
is excluded from the velocity targets. The encoded inputs are projected to
World tokens and concatenated. Within each block, masked joint attention
is followed by language cross-attention and a World-specific feed-forward
network. Output projections recover the velocity in each stream's latent
space.

\subsubsection{Action Expert.}

The Action Expert projects a noisy action chunk $U_{\tau_{\mathrm A}}$
into action tokens and conditions on $\ell$, $s$, and layer-wise World hidden
features. Let $\mathcal H_{\mathrm W}$ denote the collection of RGB and
Scalar Affordance hidden features at corresponding World layers. It contains
clean historical RGB features and evolving future RGB and Scalar Affordance
features, while excluding the Affordance Heatmap stream. The action velocity is
\begin{equation}
  \widehat v_{\mathrm A}
  =\mathcal{M}_{\mathrm A}
  \bigl(U_{\tau_{\mathrm A}},\tau_{\mathrm A},\ell,s;
  \mathcal{H}_{\mathrm W}\bigr)
  \in\mathbb{R}^{T_{\mathrm A}\times d_{\mathrm A}}.
  \label{eq:action_expert}
\end{equation}
On the joint-generation path, the features in $\mathcal H_{\mathrm W}$ evolve
with the World flow state and are exchanged within the same model evaluation.
The Action head predicts a velocity in continuous action space; integrating this velocity
yields the action chunk.

\subsubsection{Masked Joint Self-Attention.}
\label{sec:masked_joint_attention}

The Experts use independent query, key, and value projections to interact
in a common attention space at each layer. As shown in
Fig.~\ref{fig:method_overview}, RGB and Scalar Affordance tokens attend to both
streams; Affordance Heatmap tokens attend to RGB, Scalar Affordance, and
themselves; and Action tokens attend to RGB, Scalar Affordance, and themselves.
This mask blocks
Action-to-World information flow and prevents Affordance Heatmap features from
reaching Action tokens, including through intermediate RGB or Scalar Affordance
features. Affordance Heatmap supervision can still shape the shared World
features through training gradients, while RGB and Scalar Affordance provide the
World interface for action generation. Excluding Affordance Heatmap features
reduces an additional source of RGB-dominated conditioning and is intended to
mitigate RGB shortcuts, encouraging the Action Expert to make greater use of
Scalar Affordance. Direct access to RGB features remains available.

\subsection{Affordance-Aware World Representation}
\label{sec:world_representation}

We represent \emph{affordance} with two complementary components:
\textbf{Scalar Affordance} identifies task-relevant interaction regions, and
\textbf{Affordance Heatmap} associates those regions with their visual context.
Together with RGB scene prediction, they form the three future-World streams.
Throughout this paper, affordance refers to both components unless a specific
stream is named. Their encoded future latents share
a temporal grouping, enabling attention to relate interaction structure
to visual dynamics. The exact frame and latent-slot configuration is
given in Sec.~\ref{sec:supp_architecture}.

\subsubsection{RGB Future Representation.}

A frozen causal video encoder $\mathcal{E}_{\mathrm{vid}}$ compresses
the observed and future RGB sequence into history $C$ and future target
$Z_0^{\mathrm{rgb}}$:
\begin{equation}
  [C;Z_0^{\mathrm{rgb}}]
  =\mathcal E_{\mathrm{vid}}([I^-;I^+]).
  \label{eq:rgb_slots}
\end{equation}
Brackets denote temporal concatenation. Causality ensures that $C$ depends only on observed
frames, while the future slots represent appearance, object motion, and
scene-state changes. Their temporal grouping defines the alignment used
by the other two streams.

\subsubsection{Scalar Affordance Representation.}

For each future frame $t$, a scalar field
$F_t\in[0,1]^{H\times W}$ encodes the task-relevant object and its
interaction region, where $H$ and $W$ are the image height and width.
The field specifies \emph{where} to interact,
without prescribing embodiment-specific motor commands. Human and robot
videos use the same field semantics, providing a common supervision
interface across embodiments; target construction is detailed in
Sec.~\ref{sec:supp_data}.

A trainable Scalar Affordance codec, comprising encoder
$\mathcal{E}_{\mathrm{afd}}$ and decoder $\mathcal{D}_{\mathrm{afd}}$,
encodes the future fields $F^+=(F_P,\ldots,F_{P+T-1})$ as the temporally
aligned latent $Z_0^{\mathrm{afd}}=\mathcal{E}_{\mathrm{afd}}(F^+)$.
Its reconstruction $F_{\mathrm{rec}}^+=\mathcal{D}_{\mathrm{afd}}(Z_0^{\mathrm{afd}})$
is supervised by a frame-level MSE to preserve spatial affordance
information:
\begin{equation}
  \mathcal{L}_{\mathrm{rec}}
  =\mathbb{E}_{F^+}\!\left[
    \frac{1}{THW}\sum_{t=P}^{P+T-1}
    \|F_{\mathrm{rec},t}-F_t\|_{\mathrm F}^{2}
  \right],
  \label{eq:affordance_reconstruction}
\end{equation}
where $F_{\mathrm{rec},t}$ is the reconstructed field for frame $t$ and
$\|\cdot\|_{\mathrm F}$ is the Frobenius norm.
This loss supervises the encode--decode reconstruction of the clean
fields. Future Scalar Affordance generation uses latent flow matching.

\subsubsection{Affordance Heatmap Representation.}

A scalar field omits much of the appearance needed to associate an
interaction region with its object. We therefore construct a
complementary RGB-space target by rendering the field on the
corresponding frame with a fixed function $\mathcal R$:
\begin{equation}
  M_t=\mathcal R(I_t,F_t),
  \qquad
  Z_0^{\mathrm{heat}}
  =\Pi_{\mathrm{fut}}\mathcal{E}_{\mathrm{vid}}([I^-;M^+]),
  \label{eq:heatmap_renderer}
\end{equation}
where $M^+=(M_P,\ldots,M_{P+T-1})$, brackets denote temporal concatenation,
and $\Pi_{\mathrm{fut}}$ retains only the future latent slots. Using the original RGB history and the
same frozen video encoder preserves the temporal correspondence with
$Z_0^{\mathrm{rgb}}$. Affordance Heatmap prediction supplies a joint visual and
interaction target, encouraging RGB and Scalar Affordance features to retain
the information needed to reconstruct their association. The attention
mask in Fig.~\ref{fig:method_overview} prevents the Action Expert
from using Affordance Heatmap features as an additional conditioning
stream.

\subsection{Data Sources and Training Strategy}
\label{sec:data_training}

We first jointly pre-train on affordance-annotated human and robot
interactions, then adapt the Action Expert to the target robot using
vision--language--action demonstrations without affordance labels.
Human samples contain $(I^-,I^+,F^+,\ell)$; robot pre-training samples
additionally provide $(s,U_0)$. Affordance Heatmaps are derived from RGB and
Scalar Affordance targets for both sources. Dataset composition, loss weights,
and optimization settings are provided in
Secs.~\ref{sec:supp_data} and~\ref{sec:supp_optimization}.

\subsubsection{Unified Flow-Matching Objective.}

We instantiate Eq.~\ref{eq:stream_flow_loss} for each prediction branch,
using its velocity output and the conditioning information specified in
Sec.~\ref{sec:wam_architecture}. Following the path in
Eq.~\ref{eq:flow_path}, all future-World streams share flow time
$\tau_{\mathrm W}$ and receive independent Gaussian noise
$\epsilon_{\mathrm W}^m$, while actions use their own noise
$\epsilon_{\mathrm A}$ and flow time $\tau_{\mathrm A}$. The corresponding
target velocities are $u_{\mathrm W}^m=\epsilon_{\mathrm W}^m-Z_0^m$ and
$u_{\mathrm A}=\epsilon_{\mathrm A}-U_0$. The branch-specific losses are
\begin{equation}
  \begin{aligned}
    \mathcal L_m
    &=\mathbb E\!\left[
      \frac{1}{n_m}\|\widehat v_{\mathrm W}^m-u_{\mathrm W}^m\|_2^2
    \right],
    &&m\in\mathcal S_{\mathrm W},\\
    \mathcal L_{\mathrm A}
    &=\mathbb E\!\left[
      \frac{1}{n_{\mathrm A}}\|\widehat v_{\mathrm A}-u_{\mathrm A}\|_2^2
    \right].
  \end{aligned}
  \label{eq:branch_flow_losses}
\end{equation}
Here $\widehat v_{\mathrm W}^m$ is the $m$-th component of the joint World
output in Eq.~\ref{eq:world_expert}, evaluated at $\tau_{\mathrm W}$, and
$\widehat v_{\mathrm A}$ is the Action output in
Eq.~\ref{eq:action_expert}, evaluated at $\tau_{\mathrm A}$ with the
corresponding World features. The normalization $n_m$ counts the scalar
entries in $Z_0^m$, and $n_{\mathrm A}=T_{\mathrm A}d_{\mathrm A}$.
The expectations cover training samples, Gaussian noise, and flow times,
whose sampling scheme is summarized in Sec.~\ref{sec:supp_flow}.
History $C$ remains clean and is excluded from the loss. Both Scalar Affordance
and Affordance Heatmap generation are supervised by latent velocity MSE. We combine
the World prediction losses with frame-level codec reconstruction:
\begin{equation}
  \mathcal L_{\mathrm W}
  =\mathcal L_{\mathrm{rgb}}
   +\lambda_{\mathrm{afd}}\mathcal L_{\mathrm{afd}}
   +\lambda_{\mathrm{heat}}\mathcal L_{\mathrm{heat}}
   +\lambda_{\mathrm{rec}}\mathcal L_{\mathrm{rec}}.
  \label{eq:world_objective}
\end{equation}
Here the nonnegative coefficients $\lambda_{\mathrm{afd}}$,
$\lambda_{\mathrm{heat}}$, and $\lambda_{\mathrm{rec}}$ balance the auxiliary
objectives relative to RGB prediction.

\subsubsection{Stage I: Heterogeneous Joint Pre-training.}

Human videos activate only the World Expert and supervise all three
future-World streams. Robot trajectories activate both Experts and additionally
supervise action generation through the layer-wise coupling. Their
respective objectives are
\begin{equation}
  \mathcal L_{\mathrm{human}}=\mathcal L_{\mathrm W},
  \qquad
  \mathcal L_{\mathrm{robot}}
  =\mathcal L_{\mathrm W}+\lambda_{\mathrm A}\mathcal L_{\mathrm A}.
  \label{eq:heterogeneous_objective}
\end{equation}
No Action tokens or pseudo-action targets are constructed for human
videos. Human data thus improves the World representation through shared
RGB, Scalar Affordance, and Affordance Heatmap supervision, while robot data grounds that
representation in continuous control.

Training begins with World-only warmup, followed by Action pre-training with
a gradual introduction of action supervision. Robot Action training mixes
two paths: \emph{joint training}, which conditions action denoising on noisy
World features, and \emph{fixed-World training}, which conditions it on a
completed World. The curriculum progressively favors joint training and
increases timestep synchronization between the two Experts. Within the
fixed-World path, detached clean targets are gradually replaced by the
model's own predictions. This eases early optimization while preparing the
Action Expert for both joint generation and final refinement.

\subsubsection{Stage II: Task-Specific Post-training.}

We adapt to the target robot and task distribution using robot
vision--language--action demonstrations without affordance annotations.
The complete World model is frozen, and only the Action Expert is optimized
with $\mathcal L_{\mathrm A}$. Each batch retains both training paths:
noisy World features at randomly sampled timesteps maintain joint-generation
ability, while completed predicted Worlds train final Action refinement.
Gradients stop at the World interface in both paths. This preserves the
learned World representation while adapting control without restricting
the Action Expert to fixed final-World conditions. The effect of predicted-World exposure
during post-training is reported in
Table~\ref{tab:supp_architecture_ablation}.

\subsubsection{Inference.}
Given RGB history, a language instruction, and current proprioception,
we initialize future RGB, Scalar Affordance, Affordance Heatmap, and action
latent variables from independent Gaussian noise. The two Experts jointly
generate future-World latents and actions using their respective flow schedulers,
integrating from noise to data while exchanging features at each layer.
History RGB latents are re-clamped after each World update, and Action
tokens read only RGB and Scalar Affordance hidden features throughout generation.
After joint generation, we hold the completed predicted World fixed and
refine the generated action chunk through a short Action-only denoising pass.
Joint generation produces the main prediction; refinement provides final
correction under the completed World. These two inference phases correspond
to the noisy-World and fixed predicted-World training paths, respectively.
The refined action chunk is used for control. Sampling settings are provided
in Sec.~\ref{sec:supp_flow}.

\FloatBarrier
\section{Experiments}
\label{sec:experiments}

We evaluate AffordanceWAM on simulation benchmarks and real-world manipulation tasks. Our experiments are organized to address three core questions:

\begin{itemize}
    \item \textbf{Q1 (Affordance as a V-L-A Interaction Interface):} Can object-centric affordance ground language-conditioned intent in visual interaction regions and guide action generation, thereby bridging vision, language, and action?
    % Can object-centric affordance serve as an effective interaction interface that grounds language-conditioned intent in visual interaction regions and exposes this structured prediction to action generation, thereby connecting vision, language, and action?
    
    \item \textbf{Q2 (Affordance as a Human-Robot Bridge):} Does a shared affordance representation bridge human and robot interactions, enabling action-free human videos to improve robot control beyond the benefits of heterogeneous data or auxiliary supervision alone?
    % Does a shared affordance representation align human and robot interaction data, allowing action-free human videos to improve robot control rather than merely adding heterogeneous data or auxiliary prediction capacity?

    \item \textbf{Q3 (Scalable and Effective Action Learning):}
    Does the benefit of affordance-mediated human experience grow with data scale and translate into stronger manipulation performance across manipulation tasks?
\end{itemize}

\subsection{Experimental Setup}
\label{sec:exp_setup}

\paragraph{Simulation benchmarks.}
We consider RoboCasa~\cite{robocasa} and CALVIN ABC$\rightarrow$D~\cite{calvin}. RoboCasa evaluates task-specific manipulation in diverse household environments; AffordanceWAM uses 300 demonstrations per task for downstream adaptation. CALVIN ABC$\rightarrow$D evaluates cross-environment generalization by training on environments A, B, and C and testing on the unseen environment D. Following the standard protocol, we report the success rates of completing at least one to five consecutive tasks and the average completed sequence length. Unless otherwise noted, AffordanceWAM results are averaged over multiple runs with different random seeds.

% \paragraph{Baselines.}
% On RoboCasa, we compare against representative vision-language-action models, including GR00T-N1~\cite{groot_n1}, $\pi_0$~\cite{pi0}, and GR00T-N1.5~\cite{groot_n15}, as well as video and world-action models including UVA~\cite{uva}, UWM~\cite{uwm}, and Cosmos Policy~\cite{cosmos_policy}. On CALVIN, we compare against SuSIE~\cite{susie}, GR-1~\cite{gr1}, OpenVLA~\cite{openvla}, CLOVER~\cite{clover}, UniVLA~\cite{univla}, $\pi_0$~\cite{pi0}, Seer~\cite{seer}, and VPP~\cite{vpp}. Since pretraining corpora and downstream demonstration budgets differ across methods, these comparisons assess overall system-level performance; we isolate the contributions of affordance and human data through controlled ablations in Sec.~\ref{sec:ablations}.

\paragraph{Real-world setup.}
We further compare AffordanceWAM against Cosmos Policy in the real world. The evaluation contains five basic pick-and-place tasks and three complex tasks. Both methods are fine-tuned using the same 50 trajectories per task and evaluated under identical conditions over 15 trials/task. This setting evaluates whether the advantages of AffordanceWAM persist under equal-data real-world adaptation.

% \paragraph{Real-world setup.}
% Figure~\ref{fig:real-world overview} provides an overview of our real-world task suite and representative robot rollouts. The simple tasks consist of language-conditioned pick-and-place configurations that require selecting manipulated objects and target regions according to category, color, or shape, including fruit--plate placement, cup insertion, duck selection, and tool/object placement. The suite further includes three complex tasks: \emph{Close the Drawer}, \emph{Pick All the Fruits}, and \emph{Arrange the Flower}, which involve multiple interaction targets or stages of execution. We compare AffordanceWAM against Cosmos Policy using the same adaptation budget: both methods are fine-tuned with 50 trajectories per task and evaluated under identical conditions over 15 trials per task. Quantitative results are reported in Table~\ref{tab:real_world}.

\FloatBarrier
\subsection{Simulation Benchmark Results}
\label{sec:simulation_results}

\paragraph{RoboCasa.}
Table~\ref{tab:robocasa_main} reports the average success rate on RoboCasa. AffordanceWAM achieves the highest performance among the compared methods, reaching 69.4\% and exceeding Cosmos Policy by 2.3 percentage points. The result establishes the overall competitiveness of affordance-aware joint world--action modeling across diverse household manipulation tasks.
% Because the compared methods use different pretraining corpora and downstream demonstration budgets, we examine the mechanism underlying this improvement through controlled ablations in Sec.~\ref{sec:ablations}.

\begin{table}[!htbp]
  \centering
  {\small
  \setlength{\tabcolsep}{1mm}
  \begin{tabular}{@{}lcc@{}}
    \toprule
    Method & Demos/task & Average SR (\%) \\
    \midrule
    GR00T-N1~\cite{groot_n1}
      & 300 & 49.6 \\
    UVA~\cite{uva}
      & 50 & 50.0 \\
    UWM~\cite{uwm}
      & 1{,}000 & 60.8 \\
    $\pi_0$~\cite{pi0}
      & 300 & 62.5 \\
    GR00T-N1.5~\cite{groot_n15}
      & 300 & 64.1 \\
    Cosmos Policy~\cite{cosmos_policy}
      & 50 & \underline{67.1} \\
    \midrule
    \textbf{AffordanceWAM (Ours)}
      & \textbf{300} & \textbf{69.4} \\
    \bottomrule
  \end{tabular}
  }
  \caption{Results on RoboCasa. Demos/task denotes the number
  of task-specific demonstrations used.}
  \label{tab:robocasa_main}
\end{table}

\paragraph{CALVIN ABC$\rightarrow$D.}
Table~\ref{tab:calvin_main} evaluates cross-environment generalization on CALVIN ABC$\rightarrow$D. AffordanceWAM achieves the best results across all reported metrics, reaching an average sequence length of 4.22 compared with 4.01 for the strongest included baseline. It also improves the success rates for completing one to five consecutive tasks to 96.1\%, 92.8\%, 85.5\%, 77.8\%, and 69.5\%, respectively. Together with the RoboCasa results, these improvements demonstrate that the affordance-aware world--action formulation remains effective across both task-specific adaptation and unseen-environment evaluation.

\begin{table}[!htbp]
  \centering
  {\small
  \setlength{\tabcolsep}{1mm}
  \begin{tabular}{@{}lcccccc@{}}
    \toprule
    Method & 1/5 & 2/5 & 3/5 & 4/5 & 5/5
      & \shortstack{Avg.\\Len.} \\
    \midrule
    SuSIE~\cite{susie}
      & 87.0 & 69.0 & 49.0 & 38.0 & 26.0 & 2.69 \\
    GR-1~\cite{gr1}
      & 85.4 & 71.2 & 59.6 & 49.7 & 40.1 & 3.06 \\
    OpenVLA~\cite{openvla}
      & 91.3 & 77.8 & 62.0 & 52.1 & 43.5 & 3.27 \\
    CLOVER~\cite{clover}
      & \underline{96.0} & 83.5 & 70.8 & 57.5 & 45.4 & 3.53 \\
    UniVLA~\cite{univla}
      & 95.5 & 85.8 & 75.4 & 66.9 & 56.5 & 3.80 \\
    $\pi_0$~\cite{pi0}
      & 93.8 & 85.0 & 76.7 & 68.6 & 60.1 & 3.84 \\
    Seer~\cite{seer}
      & 94.4 & 87.2 & 79.9 & 72.2 & 64.3 & 3.98 \\
    VPP~\cite{vpp}
      & 95.3 & \underline{88.2} & \underline{80.3}
      & \underline{72.9} & \underline{64.5}
      & \underline{4.01} \\
    \midrule
    \shortstack[l]{\textbf{AffordanceWAM}}
      & \textbf{96.1} & \textbf{92.8} & \textbf{85.5}
      & \textbf{77.8} & \textbf{69.5} & \textbf{4.22} \\
    \bottomrule
  \end{tabular}
  }
  \caption{Results on CALVIN ABC$\rightarrow$D.
  % Columns 1/5--5/5 report the success rates (\%) of completing at least one to five consecutive tasks.
  Avg.\ Len.\ denotes the average sequence length.}
  \label{tab:calvin_main}
\end{table}

% \begin{table*}[t]
%   \centering
%   \footnotesize
%   \setlength{\tabcolsep}{3.8pt}
%   \renewcommand{\arraystretch}{0.96}
%   \begin{tabular}{@{}lcccccccccc@{}}
%     \toprule
%     Variant
%       & Human
%       & Afd.
%       & H$\rightarrow$A
%       & RoboCasa SR
%       & 1/5
%       & 2/5
%       & 3/5
%       & 4/5
%       & 5/5
%       & CALVIN Avg. Len. \\
%     \midrule

%     \textbf{AffordanceWAM}
%       & Yes & Yes & Masked
%       & \textbf{69.4}
%       & \textbf{96.1}
%       & \textbf{92.8}
%       & \textbf{85.5}
%       & \textbf{77.8}
%       & \textbf{69.5}
%       & \textbf{4.22} \\

%     w/o Human
%       & No & Yes & Masked
%       & 63.7
%       & 93.6
%       & 86.9
%       & 77.1
%       & 70.6
%       & 62.4
%       & 3.91 \\

%     w/o Human \& Afd.
%       & No & No & Masked
%       & 57.3
%       & 92.8
%       & 84.6
%       & 74.7
%       & 66.3
%       & 56.2
%       & 3.75 \\

%     w/o Afd.
%       & Yes & No & Masked
%       & 55.1
%       & 92.1
%       & 84.2
%       & 73.4
%       & 64.7
%       & 53.8
%       & 3.68 \\

%     \midrule

%     w/o H$\rightarrow$A Mask
%       & Yes & Yes & Open
%       & 63.5
%       & 93.4
%       & 85.8
%       & 76.7
%       & 69.4
%       & 59.2
%       & 3.85 \\

%     \bottomrule
%   \end{tabular}
%   \caption{
%     Controlled ablations on RoboCasa and CALVIN
%     ABC$\rightarrow$D.
%     % RC SR denotes the RoboCasa average success rate (\%); 1/5--5/5 and Avg.\ Len.\ are CALVIN metrics.
%     H$\rightarrow$A mask indicates heatmap-to-action attention is masked.
%   }
%   \label{tab:controlled_ablations}
% \end{table*}

\FloatBarrier
\subsection{Ablation Studies}
\label{sec:ablations}

\paragraph{Affordance as a Vision--Language--Action interaction interface.}
We first examine whether incorporating affordance into the generated future
World improves action learning. Here affordance comprises both Scalar
Affordance and Affordance Heatmap; w/o Afd. removes both streams and their
supervision, leaving RGB as the only World target. Under robot-only
training, affordance supervision raises RoboCasa success rate from
57.3\% to 63.7\% and CALVIN average sequence length from 3.75 to 3.91. When
human and robot videos are jointly used, the improvement becomes substantially
larger: RoboCasa success rate increases from 55.1\% to 69.4\%, while CALVIN
average sequence length increases from 3.68 to 4.22.

\begin{table}[!htbp]
  \centering
  {\small
  \setlength{\tabcolsep}{3pt}
  \begin{adjustbox}{max width=\linewidth}
  \begin{tabular}{@{}lcccccccccc@{}}
    \toprule
    Setting
      & Human
      & Afd.
      & H$\rightarrow$A
      & \shortstack{RoboCasa\\SR}
      & 1/5
      & 2/5
      & 3/5
      & 4/5
      & 5/5
      & \shortstack{CALVIN\\Avg. Len.} \\
    \midrule

    \textbf{Full}
      & Yes
      & Yes
      & Masked
      & \textbf{69.4}
      & \textbf{96.1}
      & \textbf{92.8}
      & \textbf{85.5}
      & \textbf{77.8}
      & \textbf{69.5}
      & \textbf{4.22} \\

    w/o Human
      & No
      & Yes
      & Masked
      & 63.7
      & 93.6
      & 86.9
      & 77.1
      & 70.6
      & 62.4
      & 3.91 \\

    w/o Human \& Afd.
      & No
      & No
      & Masked
      & 57.3
      & 92.8
      & 84.6
      & 74.7
      & 66.3
      & 56.2
      & 3.75 \\

    w/o Afd.
      & Yes
      & No
      & Masked
      & 55.1
      & 92.1
      & 84.2
      & 73.4
      & 64.7
      & 53.8
      & 3.68 \\

    \midrule

    w/o H$\rightarrow$A Mask
      & Yes
      & Yes
      & Open
      & 63.5
      & 93.4
      & 85.8
      & 76.7
      & 69.4
      & 59.2
      & 3.85 \\

    \bottomrule
  \end{tabular}
  \end{adjustbox}
  }
  \caption{
    Ablations on RoboCasa and CALVIN
    ABC$\rightarrow$D. 
    % RC SR denotes the RoboCasa average success
    % rate (\%). Columns 1/5--5/5 and Avg.\ Len.\ report the CALVIN
    % sequence-evaluation metrics.
    Afd. comprises both Scalar Affordance and Affordance Heatmap;
    w/o Afd. removes both streams.
    H$\rightarrow$A Mask indicates Affordance Heatmap-to-action attention is masked.
  }
  \label{tab:controlled_ablations}
\end{table}

All four controlled settings are evaluated over three independently trained
seeds. Seed-wise results and sample standard deviations are reported in
Tables~\ref{tab:supp_robocasa_seeds} and~\ref{tab:supp_calvin_seeds}.

The two affordance components are generated under language conditioning:
Scalar Affordance identifies task-relevant interaction regions, while
Affordance Heatmap associates them with visual context through auxiliary
World supervision. The Action Expert reads RGB and Scalar Affordance hidden
features to generate continuous controls. The consistent control gains
therefore support the combined affordance representation as an
effective interaction interface: language specifies task intent, affordance
grounds that intent in the visual world, and action generation determines how
the grounded interaction is executed.

\paragraph{Affordance as a Human–Robot Bridge.}
As shown in Table~\ref{tab:controlled_ablations}, the $2\times2$ factorial
ablation isolates the interaction between human-video pretraining and affordance
supervision. With affordance, adding human videos improves RoboCasa success from
63.7\% to 69.4\% and CALVIN average sequence length from 3.91 to 4.22. Without
affordance, the same human videos instead reduce the two metrics from 57.3\% to
55.1\% and from 3.75 to 3.68. The opposite trends across both benchmarks show
that additional human data alone is insufficient; its benefit depends on the
representation through which it enters the policy. This interaction amounts to 7.9 percentage points on RoboCasa and 0.38 average
sequence length on CALVIN. Without a shared interaction target, human videos
introduce substantial appearance and embodiment variation without clarifying
which changes are relevant to control. Affordance supervision instead expresses
human and robot interactions through shared Scalar Affordance and Affordance
Heatmap targets, allowing
both domains to supervise task-relevant World dynamics. Since human clips never
activate the Action Expert, their control gains must arise through the learned
World representation and its coupling to action generation. The sign reversal
therefore provides direct evidence that affordance is the mechanism that makes
action-free human experience useful for robot learning.

% \begin{table}[t]
%   \centering
%   {\small
%   \setlength{\tabcolsep}{1mm}
%   \begin{tabular}{cccc}
%     \toprule
%     Human & Afd. & RoboCasa SR (\%) & CALVIN Len. \\
%     \midrule
%     Yes & No  & 55.1 & 3.68 \\
%     No  & No  & 57.3 & 3.75 \\
%     No  & Yes & 63.7 & 3.91 \\
%     Yes & Yes & \textbf{69.4} & \textbf{4.22} \\
%     \bottomrule
%   \end{tabular}
%   }
%   \caption{Factorial ablation of human-video pretraining and affordance supervision. Human data improves robot control only when the two embodiments are connected through a shared affordance target.}
%   \label{tab:human_affordance_ablation}
% \end{table}

\paragraph{Complementary roles of Scalar Affordance and Affordance Heatmap.}
We additionally remove either the Scalar Affordance stream or the
Affordance Heatmap stream, separately from w/o Afd., which removes both.
Qualitative predictions reveal two distinct failure modes. Without Scalar
Affordance, the RGB-dominated Affordance Heatmap permits the World Expert
to model visual appearance while largely discarding interaction-specific
structure. Without Affordance Heatmap, the predicted Scalar Affordance
increasingly drifts from its corresponding object regions, indicating weaker
consistency between visual dynamics and interaction prediction. Removing
Affordance Heatmap also reduces RoboCasa success from 69.4\% to 65.4\% and
CALVIN average sequence length from 4.22 to 3.99
(Table~\ref{tab:supp_architecture_ablation}).

The two streams therefore provide complementary inductive biases. Scalar
Affordance explicitly represents task-relevant interaction regions, whereas
Affordance Heatmap encourages these regions to remain spatially aligned with
the predicted visual world. Their distinct failure modes suggest that the
benefit arises from the structured roles of the two representations, rather
than treating additional prediction streams as interchangeable capacity.
Qualitative examples are provided in Figure~\ref{fig:supp_qualitative}.

% \paragraph{Complementary roles of Scalar Affordance and Affordance Heatmap.}
% Scalar Affordance explicitly represents task-relevant interaction regions,
% while Affordance Heatmap encourages their association with visual context.
% Removing only Affordance Heatmap reduces RoboCasa success from 69.4\% to
% 65.4\% and CALVIN average sequence length from 4.22 to 3.99
% (Table~\ref{tab:supp_architecture_ablation}), supporting its contribution
% alongside Scalar Affordance. This component ablation differs from w/o Afd.,
% which removes both affordance streams. The qualitative Affordance Heatmap
% predictions in Figure~\ref{fig:supp_qualitative} illustrate spatial drift,
% RGB-copy collapse, and wrong-object localization as failure modes of
% interaction prediction.

\paragraph{Scaling human interaction experience.}
We vary the fraction of affordance-annotated human data while keeping robot-data exposure, training updates, and all other settings fixed. As shown in Table~\ref{tab:human_scaling}, RoboCasa success increases monotonically but non-uniformly: the first 60\% of human data yields only a 1.5-point gain, whereas scaling from 60\% to 100\% contributes a further 4.2 points, accounting for approximately 74\% of the total improvement.
Seed-wise values and sample standard deviations for each data fraction are
provided in Table~\ref{tab:supp_human_scaling}.

% This threshold-like trend suggests that human data becomes substantially more useful once it provides sufficient coverage of diverse object interactions, allowing the shared affordance representation to better bridge human and robot experience. Because robot action supervision remains fixed, the resulting gain reflects more effective use of the same action-labeled robot dataset, supporting affordance-mediated human video as a practical route toward more data-efficient action learning.

% \paragraph{Scaling human interaction experience.}
% We next examine how the benefit changes with the amount of human interaction experience. We vary the fraction of the affordance-annotated human dataset while keeping robot-data exposure, training updates, and all other settings fixed. As shown in Table~\ref{tab:human_scaling}, performance improves monotonically but markedly non-uniformly.

% Increasing the human-data fraction from 0\% to 60\% raises RoboCasa success rate by only 1.5 percentage points, from 63.7\% to 65.2\%. In contrast, increasing it from 60\% to 100\% produces a further 4.2-point gain, accounting for approximately 74\% of the total improvement. The marginal gains also increase with scale, from $+0.2$, $+0.4$, and $+0.9$ points in the first three intervals to $+1.9$ and $+2.3$ points beyond 60\%.

This threshold-like pattern suggests that limited human data provides insufficient interaction coverage to overcome the substantial visual and embodiment gap. Once the human dataset reaches broader coverage of objects and interaction regions, additional examples can reinforce the shared affordance structure instead of merely adding appearance variation. Since robot action supervision and optimization updates remain fixed throughout this comparison, the improvement reflects more effective use of the same action-labeled robot dataset. The result supports affordance-mediated human interaction experience as a practical route toward more data-efficient robot action learning.

\begin{table}[!htbp]
  \centering
  {\small
  \setlength{\tabcolsep}{1mm}
  \begin{tabular}{lcccccc}
    \toprule
    Human data (\%) & 0 & 20 & 40 & 60 & 80 & 100 \\
    \midrule
    RoboCasa SR (\%) & 63.7 & 63.9 & 64.3 & 65.2 & 67.1 & \textbf{69.4} \\
    \bottomrule
  \end{tabular}
  }
  \caption{Scaling affordance-annotated human video on RoboCasa. Robot-data exposure and training updates are fixed.}
  \label{tab:human_scaling}
\end{table}

\FloatBarrier
\paragraph{Effect of Affordance Heatmap-to-action masking.}
Affordance Heatmap retains substantial RGB appearance. Allowing the Action
Expert to read its features introduces an additional RGB-rich pathway that
may favor appearance-based shortcuts. Masking this pathway reduces the
additional RGB signal and is intended to encourage greater use of Scalar
Affordance. It mitigates this potential shortcut while preserving direct
access to RGB features; it does not require actions to depend exclusively on
Scalar Affordance. Opening the Affordance Heatmap-to-action attention path
reduces RoboCasa success from 69.4\% to 63.5\% and CALVIN average sequence
length from 4.22 to 3.85 (Table~\ref{tab:supp_architecture_ablation}).
These results support keeping Affordance Heatmap as auxiliary World
supervision and using RGB and Scalar Affordance as the direct World
interface for control, consistent with the intended reduction of RGB shortcuts.

\FloatBarrier
\subsection{Real-World Evaluation}
\label{sec:real_world}

\begin{figure}[!htbp]
\centering
\includegraphics[width=0.95\textwidth]{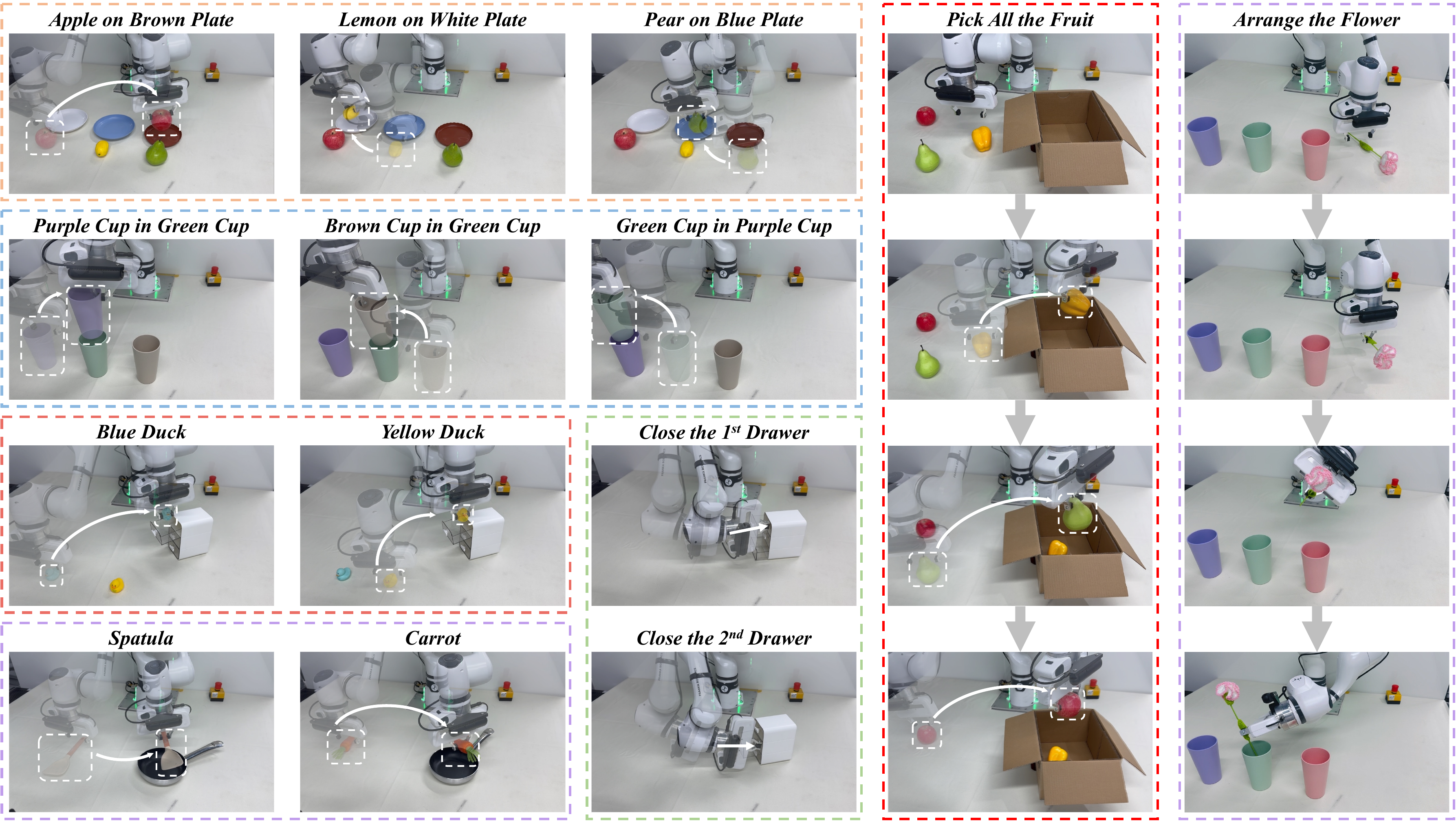}
\caption{Overview of the real-world manipulation tasks. The task suite contains simple language-conditioned manipulation tasks involving fruit--plate placement, cup insertion, object selection, and tool/object placement (left), together with three complex tasks: \emph{Close the Drawer}, \emph{Pick All the Fruit}, and \emph{Arrange the Flower} (right). 
% White dashed boxes mark manipulated objects or target regions, white arrows indicate the intended object motion, and gray arrows show the temporal progression of representative rollouts. 
% The simple tasks evaluate manipulation with respect to object category, color, and shape, whereas the complex tasks require multiple interaction targets or stages of execution.
}
\label{fig:real-world_overview}
\end{figure}

Figure~\ref{fig:real-world_overview} provides an overview of our real-world task suite and representative robot rollouts, comprising five simple language-conditioned tasks that test category-, color-, and shape-based object and target selection, as well as three complex tasks: \emph{Close the Drawer}, \emph{Pick All the Fruits}, and \emph{Arrange the Flower}.

\begin{table}[!htbp]
  \centering
  {\small
  \setlength{\tabcolsep}{1mm}
  \begin{tabular}{@{}lcccccc@{}}
    \toprule

    \multicolumn{1}{@{}l}{\textit{(a) Basic real-world tasks}}
      & \multicolumn{1}{c}{Category}
      & \multicolumn{2}{c}{Color}
      & \multicolumn{2}{c}{Shape}
      & \\
    \cmidrule(lr){2-2}
    \cmidrule(lr){3-4}
    \cmidrule(lr){5-6}
    \textbf{Method}
      & Fruit to Plate
      & Cup
      & Duck
      & Carrot
      & Spatula
      & Avg. \\
    \midrule

    Cosmos Policy~\cite{cosmos_policy}
      & 60.0
      & 40.0
      & 46.7
      & 40.0
      & 26.7
      & 42.7 \\
    \textbf{AffordanceWAM (Ours)}
      & \textbf{86.7}
      & \textbf{60.0}
      & \textbf{73.3}
      & \textbf{86.7}
      & \textbf{66.7}
      & \textbf{74.7} \\

    \midrule

    \multicolumn{1}{@{}l}{\textit{(b) Complex real-world tasks}}
      & \multicolumn{1}{c}{Close the Drawer}
      & \multicolumn{3}{c}{Pick All the Fruit}
      & \multicolumn{2}{c}{Arrange the Flower} \\
    \cmidrule(lr){2-2}
    \cmidrule(lr){3-5}
    \cmidrule(lr){6-7}
    \textbf{Method}
      & Success
      & $\geq$1 Fruit
      & $\geq$2 Fruits
      & $\geq$3 Fruits
      & \multicolumn{2}{c}{Success} \\
    \midrule

    Cosmos Policy~\cite{cosmos_policy}
      & 46.7
      & 73.3
      & 53.3
      & 26.7
      & \multicolumn{2}{c}{6.7} \\
    \textbf{AffordanceWAM (Ours)}
      & \textbf{80.0}
      & \textbf{86.7}
      & \textbf{73.3}
      & \textbf{53.3}
      & \multicolumn{2}{c}{\textbf{20.0}} \\

    \bottomrule
  \end{tabular}
  }
  \caption{Success rates (\%) on real-world manipulation tasks. Both
  methods use 50 fine-tuning trajectories per task and are evaluated
  over 15 trials under identical conditions. Panel (a) reports results
  on basic tasks and their macro-average. Panel (b) reports results on
  \emph{Close the Drawer} and \emph{Arrange the Flower}, together with
  stage-wise success rates on \emph{Pick All the Fruit}.}
  \label{tab:real_world}
\end{table}

Under the same adaptation budget, AffordanceWAM outperforms Cosmos Policy on all simple tasks and raises the macro-average success rate from 42.7\% to 74.7\% (Table~\ref{tab:real_world}). It also improves drawer-closing success from 46.7\% to 80.0\%, improves all three fruit-placement thresholds---including the three-fruit success rate from 26.7\% to 53.3\%---and raises flower-arrangement success from 6.7\% to 20.0\%. These results extend the simulation findings to physical manipulation and show that the affordance-aware interaction representation remains beneficial under equal-data real-world adaptation.

\FloatBarrier
\begin{samepage}
\section{Conclusion}

In this work, we present AffordanceWAM to bridge the gap between human and
robot embodiments. Rather than aligning human motion
with robot actions or modeling the future with RGB imagery alone, AffordanceWAM
forecasts Scalar Affordance and Affordance Heatmap as a shared interaction
representation within the future World across embodiments. A directionally coupled dual-expert architecture learns
this World from human and robot data while grounding action generation
exclusively in robot trajectories. AffordanceWAM achieves strong results on
RoboCasa, CALVIN, and real-world tasks.
% Controlled ablations further show that
% human video improves robot control only when affordance provides the shared
% prediction. 
Controlled ablations further suggest that the benefits of human video for robot control are pronounced when affordance serves as the shared predictive representation across embodiments.
\par
\end{samepage}

% ---------------------------------------------------------------
% Bibliography
% Let references follow the conclusion to avoid a nearly empty overflow page.
\FloatBarrier
\bibliographystyle{assets/plainnat}
\bibliography{main}

% ---------------------------------------------------------------
% Appendix
\clearpage
\beginappendix
% Supplementary material for AffordanceWAM.
\setcounter{secnumdepth}{2}

This supplement provides data and implementation details, training and
sampling settings, seed-wise results, and additional quantitative and
qualitative analyses. Notation and model objectives follow
Sec.~\ref{sec:method}.

\FloatBarrier
\section{Data Preparation}
\label{sec:supp_data}

\FloatBarrier
\subsection{Sources and Splits}

Human pre-training videos come from Ego4D-FHO, EPIC-KITCHENS VISOR, HOI4D,
H2O, and MECCANO. Robot trajectories come from RoboInter (DROID and RH20T),
BridgeData V2, InternData-A1, and RoboCasa365.
Table~\ref{tab:supp_data_composition} summarizes the retained scale,
sampling ratio, and downstream data budgets. The stratified validation
holdout is used for training-time validation and checkpoint selection.

\begin{table}[!htbp]
  \centering
  {\small
  \setlength{\tabcolsep}{1.2mm}
  \begin{tabular}{@{}p{0.34\columnwidth}p{0.60\columnwidth}@{}}
    \toprule
    Scope & Scale or configuration \\
    \midrule
    Stage-I human & More than 300 hours \\
    Stage-I robot & Approximately 0.4 million trajectories \\
    Stage-I sampling & Human:robot \(=1{:}1\) \\
    Validation & Fixed 2,000 clips plus a stratified 1\% holdout \\
    RoboCasa Stage II & \(24\times300=7{,}200\) demonstrations \\
    CALVIN Stage II & Complete ABC training split \\
    Real Stage II & \(8\times50=400\) trajectories \\
    \bottomrule
  \end{tabular}
  }
  \caption{Data composition for heterogeneous pre-training, validation, and
  task-specific post-training.}
  \label{tab:supp_data_composition}
\end{table}

\FloatBarrier
\subsection{Affordance Target Construction}

We construct targets from the source datasets' existing object and
interaction annotations. The following
procedure implements the shared field semantics defined in
Sec.~\ref{sec:world_representation}.

\begin{enumerate}
  \item \textbf{Clip and annotation alignment.} Sample clips using the
  temporal and spatial configuration in
  Table~\ref{tab:supp_model_configuration}, and align object and
  interaction-region annotations with the sampled timestamps.
  \item \textbf{Label normalization.} Map source-specific annotations to
  the common task-relevant object and interaction-region labels.
  \item \textbf{Field rasterization.} Construct $F_t$ using a Gaussian
  kernel and per-frame peak normalization. Interpolate missing annotations
  over at most two frames; under occlusion, retain only visible support.
  \item \textbf{Affordance Heatmap targets.} Apply the fixed renderer in
  Eq.~\ref{eq:heatmap_renderer} to each aligned RGB--Scalar Affordance pair.
\end{enumerate}

\FloatBarrier
\subsection{Filtering and Quality Control}

Human clips are filtered for excessive camera shake using motion derived
from camera parameters and for invalid Affordance Heatmaps under the central-$9/16$
region criterion. Retained clips undergo random manual review. Robot
trajectories are checked for temporal alignment, valid states and actions,
and valid interaction annotations. These checks precede the retained
counts in Table~\ref{tab:supp_data_composition}.

\FloatBarrier
\section{Model Implementation}
\label{sec:supp_architecture}

\FloatBarrier
\subsection{Backbone and Expert Dimensions}

Table~\ref{tab:supp_model_configuration} specifies the pretrained backbone,
Expert dimensions, and prediction interface. Native hidden widths and
joint-attention dimensions are listed separately, following the
architecture in Sec.~\ref{sec:wam_architecture}.

\begin{table}[!htbp]
  \centering
  {\small
  \setlength{\tabcolsep}{1.2mm}
  \begin{tabular}{@{}p{0.30\columnwidth}p{0.64\columnwidth}@{}}
    \toprule
    Component & Configuration \\
    \midrule
    World initialization & Wan2.2-TI2V-5B \\
    World Expert & 30 layers; hidden dimension 3,072 \\
    Action Expert & 30 layers; hidden dimension 1,024; FFN dimension 4,096 \\
    Joint attention & 24 heads \(\times\) 128 dimensions \(=3{,}072\) \\
    Training clip & 17 frames: 5 observed and 12 future; 10 FPS; \(256\times320\) \\
    Deployment history & 5 observed RGB frames \\
    Action output & \(12\times7\) continuous values \\
    \bottomrule
  \end{tabular}
  }
  \caption{Model dimensions and input/output configuration.}
  \label{tab:supp_model_configuration}
\end{table}

\FloatBarrier
\subsection{Temporal Latent Layout}

The configuration corresponds to $P=5$, $T=T_{\mathrm A}=12$, and
$d_{\mathrm A}=7$ in the Method. The causal video encoder produces five
temporal slots: one for frame 0, one for frames 1--4, and three for the
future groups 5--8, 9--12, and 13--16. The first two slots form $C$;
the remaining three form $Z_0^{\mathrm{rgb}}$. The Scalar Affordance codec
produces three slots with the same future grouping, and the Affordance Heatmap
encoder output is restricted to these three future slots. Each slot
contains spatial latent features, not a single Transformer token.

\FloatBarrier
\section{Training Configuration}
\label{sec:supp_optimization}

Stage I calibrates the Scalar Affordance codec and warms up the World
Expert before introducing Action training. World-only warmup uses 5,000
updates for the 8-slot configuration and 7,000 for the 11-slot configuration.
During Action pre-training, the action-loss weight increases gradually from
zero to one. Fixed-World conditioning is emphasized early, then reduced as joint
noisy-World/Action training becomes dominant. Within the joint path,
synchronized timesteps become more frequent; within the fixed-World path,
clean target Worlds are gradually replaced entirely by predicted Worlds.
This curriculum prepares joint generation and final refinement together.

The final loss weights are $\lambda_{\mathrm{afd}}=0.2$,
$\lambda_{\mathrm{heat}}=0.5$, $\lambda_{\mathrm{rec}}=0.1$, and
$\lambda_{\mathrm A}=1$, with unit weight on RGB prediction.
Here $\mathrm{afd}$ denotes the Scalar Affordance branch and
$\mathrm{heat}$ denotes the Affordance Heatmap branch.
We use AdamW and BF16 training on 64 NVIDIA H200 GPUs.

Stage II freezes the complete World path and trains only the Action Expert
for 10,000 updates at a learning rate of $10^{-5}$ for each target benchmark.
Within each batch, 75\% of samples use frozen World features at randomly
sampled noise timesteps to preserve joint-generation ability; the remaining
25\% use completed predicted Worlds to train final Action refinement.
Both paths stop gradients at the World interface.

\FloatBarrier
\section{Joint Generation and Action Refinement}
\label{sec:supp_flow}

Joint training uses noisy World and Action latents along the flow path in
Eq.~\ref{eq:flow_path}. World timesteps are sampled from
$\mathcal U(0,1)$; Action timesteps are either synchronized with World
or sampled independently. Fixed-World training holds the completed World
condition fixed while denoising actions. During pre-training this condition
transitions from detached clean targets to model predictions; Stage II uses
completed predicted Worlds for its fixed-World samples. In both paths,
Action reads RGB and Scalar Affordance features, with Affordance Heatmap
features excluded by the attention mask.

All main AffordanceWAM policy results use the same inference procedure:
\textbf{35 World steps and 35 Action steps of joint generation, followed by
8 steps of Action refinement with strength 0.25}. Refinement operates on
the jointly generated action chunk while keeping the completed predicted
World fixed. Thus, noisy-World training supports main generation, and
fixed predicted-World training supports final correction.

\FloatBarrier
\section{Evaluation Protocols and Seed-Wise Results}
\label{sec:supp_statistics}

\FloatBarrier
\subsection{Evaluation Units and Aggregation}

Controlled RoboCasa and CALVIN experiments use three independently trained
seeds. Each RoboCasa seed is evaluated on 24 tasks with 50 episodes per
task, giving 1,200 episodes per seed and 3,600 per setting. Each CALVIN
seed is evaluated on 1,000 five-task sequences, giving 3,000 per setting.
We report the mean and sample standard deviation across training seeds;
the seed is the unit of run-to-run variability.

Tables~\ref{tab:supp_robocasa_seeds} and~\ref{tab:supp_calvin_seeds}
expand the aggregate results in Table~\ref{tab:controlled_ablations}.
In both tables, w/o Afd. removes both Scalar Affordance and Affordance
Heatmap, leaving an RGB-only World target; w/o Human \& Afd. additionally
removes human pre-training data.
Seed identifiers are aligned across the controlled settings.

\begin{table}[!htbp]
  \centering
  {\small
  \setlength{\tabcolsep}{1.7mm}
  \begin{tabular}{lrrrr}
    \toprule
    Setting & Seed 1 & Seed 2 & Seed 3 & Mean \(\pm\) std \\
    \midrule
    Full & 69.9 & 69.4 & 68.9 & \textbf{69.4 \(\pm\) 0.5} \\
    w/o Human & 63.8 & 63.2 & 64.1 & 63.7 \(\pm\) 0.5 \\
    w/o Human \& Afd. & 57.4 & 56.8 & 57.7 & 57.3 \(\pm\) 0.5 \\
    w/o Afd. & 55.2 & 54.5 & 55.6 & 55.1 \(\pm\) 0.6 \\
    \bottomrule
  \end{tabular}
  }
  \caption{Seed-wise controlled ablations on RoboCasa. Values are average
  task success rates in percent. Each seed contains 1,200 evaluation episodes;
  the final column reports the mean and sample standard deviation over three
  independently trained seeds.}
  \label{tab:supp_robocasa_seeds}
\end{table}

\begin{table}[!htbp]
  \centering
  {\small
  \setlength{\tabcolsep}{1.8mm}
  \begin{tabular}{llrrrrrr}
    \toprule
    Setting & Run & 1/5 (\%) & 2/5 (\%) & 3/5 (\%) & 4/5 (\%) & 5/5 (\%) & Avg. Len. \\
    \midrule
    \multirow{4}{*}{Full}
        & Seed 1 & 96.4 & 92.8 & 85.5 & 78.6 & 69.5 & 4.23 \\
        & Seed 2 & 95.8 & 93.4 & 84.8 & 77.0 & 70.5 & 4.22 \\
        & Seed 3 & 96.1 & 92.2 & 86.2 & 77.8 & 68.5 & 4.21 \\
        & Mean $\pm$ std
        & \textbf{96.1 $\pm$ 0.3}
        & \textbf{92.8 $\pm$ 0.6}
        & \textbf{85.5 $\pm$ 0.7}
        & \textbf{77.8 $\pm$ 0.8}
        & \textbf{69.5 $\pm$ 1.0}
        & \textbf{4.22 $\pm$ 0.01} \\
    \midrule
    \multirow{4}{*}{w/o Human}
        & Seed 1 & 93.6 & 87.5 & 76.4 & 70.6 & 63.1 & 3.91 \\
        & Seed 2 & 94.1 & 86.3 & 77.8 & 69.9 & 62.4 & 3.91 \\
        & Seed 3 & 93.1 & 86.9 & 77.1 & 71.3 & 61.7 & 3.90 \\
        & Mean $\pm$ std
        & 93.6 $\pm$ 0.5
        & 86.9 $\pm$ 0.6
        & 77.1 $\pm$ 0.7
        & 70.6 $\pm$ 0.7
        & 62.4 $\pm$ 0.7
        & 3.91 $\pm$ 0.01 \\
    \midrule
    \multirow{4}{*}{w/o Human \& Afd.}
        & Seed 1 & 93.3 & 84.6 & 75.4 & 66.3 & 56.2 & 3.76 \\
        & Seed 2 & 92.3 & 85.2 & 74.7 & 65.6 & 56.9 & 3.75 \\
        & Seed 3 & 92.8 & 84.0 & 74.0 & 67.0 & 55.5 & 3.73 \\
        & Mean $\pm$ std
        & 92.8 $\pm$ 0.5
        & 84.6 $\pm$ 0.6
        & 74.7 $\pm$ 0.7
        & 66.3 $\pm$ 0.7
        & 56.2 $\pm$ 0.7
        & 3.75 $\pm$ 0.01 \\
    \midrule
    \multirow{4}{*}{w/o Afd.}
        & Seed 1 & 92.6 & 84.2 & 73.4 & 64.0 & 54.5 & 3.69 \\
        & Seed 2 & 92.1 & 83.6 & 74.1 & 65.4 & 53.1 & 3.68 \\
        & Seed 3 & 91.6 & 84.8 & 72.7 & 64.7 & 53.8 & 3.68 \\
        & Mean $\pm$ std
        & 92.1 $\pm$ 0.5
        & 84.2 $\pm$ 0.6
        & 73.4 $\pm$ 0.7
        & 64.7 $\pm$ 0.7
        & 53.8 $\pm$ 0.7
        & 3.68 $\pm$ 0.01 \\
    \bottomrule
  \end{tabular}
  }
  \caption{Seed-wise controlled ablations on CALVIN ABC\(\rightarrow\)D.
  Each seed contains 1,000 five-task evaluation sequences. Values after
  \(\pm\) are sample standard deviations over three independently trained
  seeds.}
  \label{tab:supp_calvin_seeds}
\end{table}

\FloatBarrier
\subsection{Human-Data Scaling}

Table~\ref{tab:supp_human_scaling} provides the three individual runs and
sample standard deviations underlying Table~\ref{tab:human_scaling}.
The experimental controls follow Sec.~\ref{sec:ablations}.

\begin{table}[!htbp]
  \centering
  {\small
  \setlength{\tabcolsep}{1.1mm}
  \begin{tabular}{lrrrrrr}
    \toprule
    Human data (\%) & 0 & 20 & 40 & 60 & 80 & 100 \\
    \midrule
        Seed 1 & 63.8 & 64.4 & 64.3 & 65.7 & 67.1 & 69.9 \\
        Seed 2 & 63.2 & 63.4 & 63.8 & 64.7 & 67.7 & 69.4 \\
        Seed 3 & 64.1 & 63.9 & 64.8 & 65.2 & 66.5 & 68.9 \\
        Mean & 63.7 & 63.9 & 64.3 & 65.2 & 67.1 & \textbf{69.4} \\
        Std & 0.5 & 0.5 & 0.5 & 0.5 & 0.6 & 0.5 \\
    \bottomrule
  \end{tabular}
  }
  \caption{Seed-wise RoboCasa success rates for the human-data scaling
  experiment. The standard deviation is computed across three training
  seeds.}
  \label{tab:supp_human_scaling}
\end{table}

\FloatBarrier
\section{Additional Quantitative Evaluations}
\label{sec:supp_additional_ablations}

\FloatBarrier
\subsection{Additional Architectural Ablations}

Table~\ref{tab:supp_architecture_ablation} extends the controls in
Table~\ref{tab:controlled_ablations} with alternative fusion designs,
removal of predicted-World post-training, and removal of the Affordance Heatmap
stream. Data and update budgets are held fixed. The fusion and masking
controls also match token count and trainable parameters.

\begin{table}[!htbp]
  \centering
  {\small
  \setlength{\tabcolsep}{1.4mm}
  \begin{tabular}{lrr}
    \toprule
    Variant & RoboCasa SR & CALVIN Len. \\
    \midrule
    Masked Joint Self-Attn. & \textbf{69.4 \(\pm\) 0.5} & \textbf{4.22 \(\pm\) 0.01} \\
    Late fusion & 65.9 \(\pm\) 0.6 & 4.00 \(\pm\) 0.01 \\
    Independent World\(\rightarrow\)Action & 64.5 \(\pm\) 0.6 & 3.90 \(\pm\) 0.01 \\
    Open Affordance Heatmap\(\rightarrow\)Action & 63.5 \(\pm\) 0.6 & 3.85 \(\pm\) 0.01 \\
    w/o predicted-World post-training & 66.6 \(\pm\) 0.6 & 4.06 \(\pm\) 0.01 \\
    w/o Affordance Heatmap stream & 65.4 \(\pm\) 0.6 & 3.99 \(\pm\) 0.01 \\
    \bottomrule
  \end{tabular}
  }
  \caption{Architectural ablations under the same data and update budget.
  Fusion and masking controls additionally match token count and trainable
  parameters; the stream-removal row is a targeted component ablation.}
  \label{tab:supp_architecture_ablation}
\end{table}

\paragraph{Variant definitions.}
In \emph{late fusion}, the Experts communicate only through a final
World-to-Action fusion module. The \emph{independent pipeline} first
completes World prediction, then supplies it as a fixed condition to a
separately executed Action Expert. \emph{Open Affordance Heatmap$\rightarrow$Action}
removes the block on Affordance Heatmap keys for Action queries. The
\emph{predicted-World post-training} ablation removes Stage II exposure
to World predictions; the \emph{Affordance Heatmap} ablation removes that
auxiliary stream. The post-training ablation concerns training and is
separate from the final inference-time Action refinement in
Sec.~\ref{sec:supp_flow}.

\FloatBarrier
\section{Qualitative Analysis}
\label{sec:supp_diagnostics}

\FloatBarrier
\subsection{Qualitative Predictions and Failure Modes}

\begin{figure}[!htbp]
  \centering
  \includegraphics[width=1.0\textwidth]{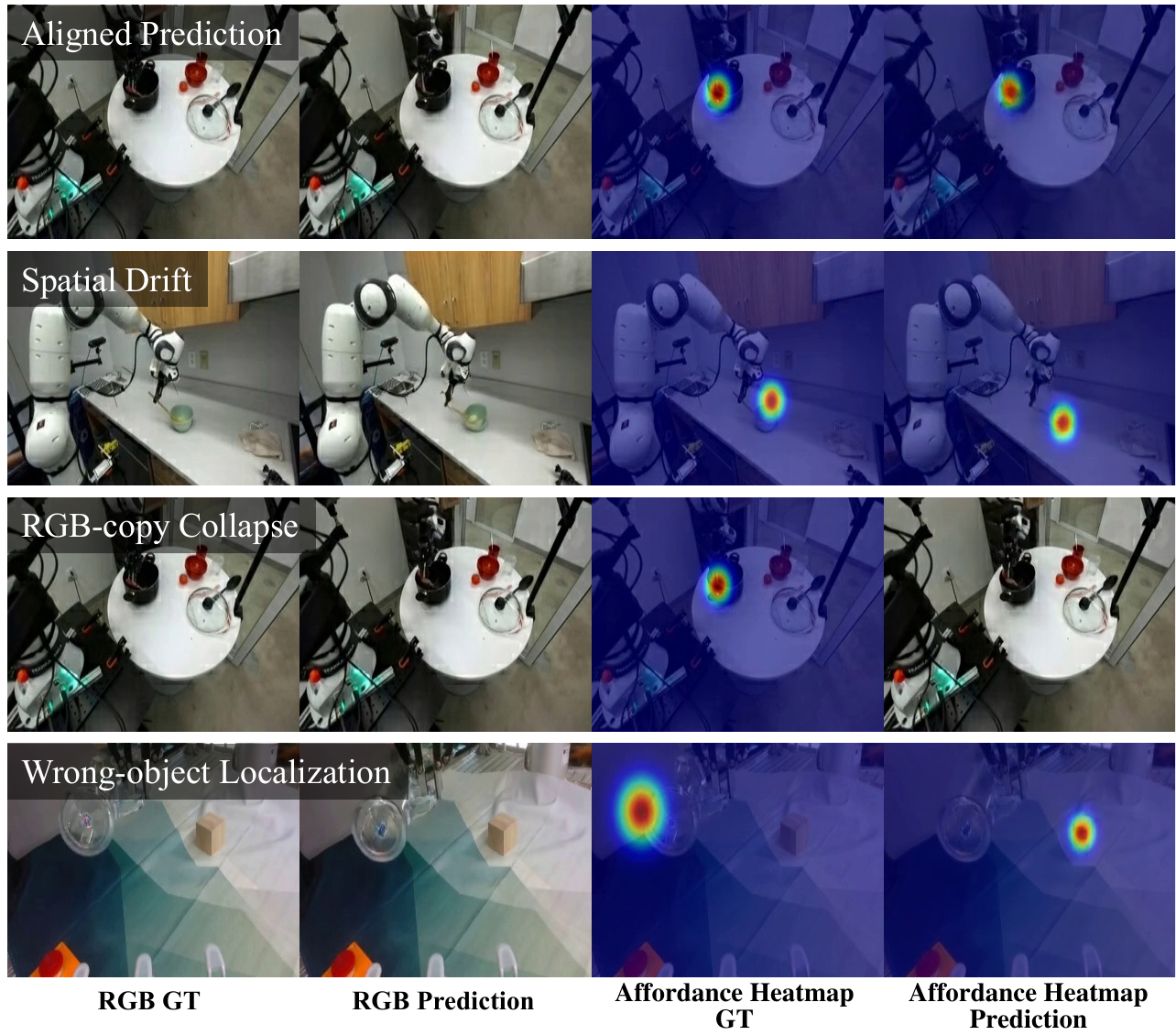}
  \caption{Qualitative RGB and Affordance Heatmap predictions.
  From left to right: RGB GT, RGB Prediction, Affordance Heatmap GT,
  and Affordance Heatmap Prediction, where GT denotes ground truth.
  From top to bottom: aligned prediction, spatial drift, RGB-copy collapse,
  and wrong-object localization.}
  \label{fig:supp_qualitative}
\end{figure}

Figure~\ref{fig:supp_qualitative} compares generated RGB observations and
Affordance Heatmaps with their clean targets. The aligned example localizes the interaction
region on the intended object. The remaining rows illustrate three
failure types:
\begin{itemize}
  \item \textbf{Spatial drift:} a predicted interaction region is displaced
  from the corresponding object despite plausible RGB prediction.
  \item \textbf{RGB-copy collapse:} appearance structure dominates the
  Affordance Heatmap prediction, weakening interaction-specific localization.
  \item \textbf{Wrong-object localization:} a compact prediction selects
  an object inconsistent with the instruction.
\end{itemize}

Further diagnostic categories are \emph{correct World, failed control},
where plausible visual predictions are followed by contact, kinematic, or
execution errors, and \emph{long-horizon degradation}, where localization
quality worsens across successive future groups. These categories separate
representation errors from downstream control failures.

\end{document}